\documentclass[10pt,twocolumn,letterpaper]{article}

\usepackage[pagenumbers]{cvpr} 

\usepackage{graphicx}
\usepackage{amsmath}
\usepackage{amssymb}
\usepackage{booktabs}
\usepackage{times}
\usepackage{epsfig}
\usepackage{graphicx}
\usepackage{multirow}
\usepackage[table,xcdraw]{xcolor}
\usepackage[pagebackref,breaklinks,colorlinks]{hyperref}
\usepackage[capitalize]{cleveref}
\crefname{section}{Sec.}{Secs.}
\Crefname{section}{Section}{Sections}
\Crefname{table}{Table}{Tables}
\crefname{table}{Tab.}{Tabs.}

\def\cvprPaperID{43} 
\def\confName{CVPR}
\def\confYear{2022}

\begin{document}

\title{ViTOL: Vision Transformer for Weakly Supervised Object Localization}

\author{Saurav Gupta \hspace{1cm} Sourav Lakhotia \hspace{1cm} Abhay Rawat \hspace{1cm} Rahul Tallamraju\\
Mercedes-Benz Research and Development India\\
{\tt\small \{saurav.gupta, sourav.lakhotia, abhay.rawat, rahul.tallamraju\}@mercedes-benz.com}
}

\maketitle

\begin{abstract}
   Weakly supervised object localization (WSOL) aims at predicting object locations in an image using only image-level category labels. 
   Common challenges that image classification models encounter when localizing objects are, (a) they tend to look at the most discriminative features in an image that confines the localization map to a very small region, (b) the localization maps are class agnostic, and the models highlight objects of multiple classes in the same image and, (c) the localization performance is affected by background noise. To alleviate the above challenges we introduce the following simple changes through our proposed method \textbf{ViTOL}. We leverage the vision-based transformer for self-attention and introduce a patch-based attention dropout layer ({p-ADL}) to increase the coverage of the localization map and a gradient attention rollout mechanism to generate class-dependent attention maps. 
We conduct extensive quantitative, qualitative and ablation experiments on the ImageNet-1K and CUB datasets. We achieve state-of-the-art  MaxBoxAcc-V2 localization scores of $70.47\%$ and $73.17\%$ on the two datasets respectively. Code is available on \href{https://github.com/Saurav-31/ViTOL}{https://github.com/Saurav-31/ViTOL}.

\end{abstract}

\begin{figure}[t]
	\begin{center}
		\includegraphics[width=1\linewidth, trim = 1.5cm 3cm 1.5cm 0, clip]{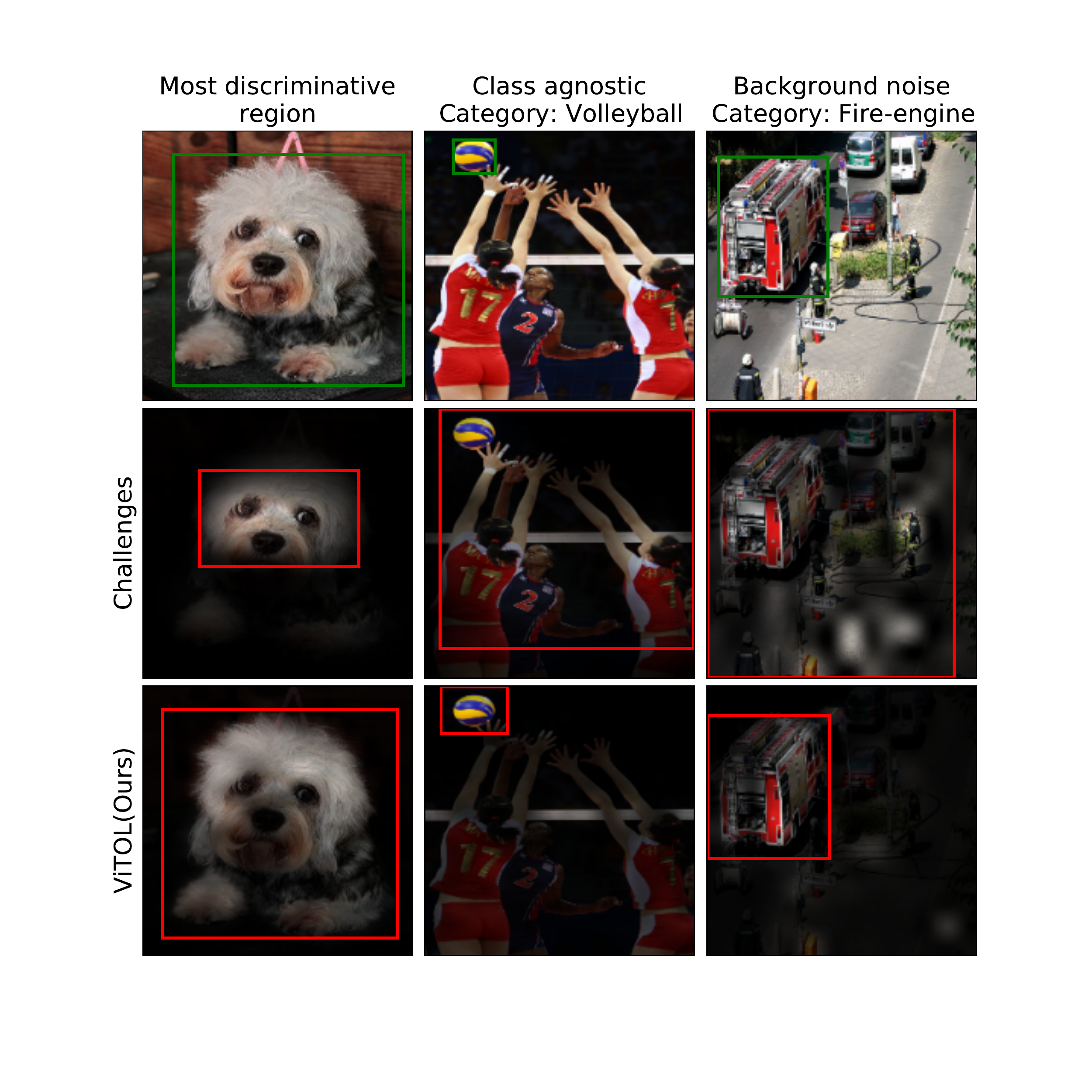}
	\end{center}
	\caption{\textbf{Common Challenges in WSOL (row2) and ViTOL output (row3)}  are visualized for three \textbf{query images (row1)}. The GT box is shown in green and the predicted box is shown in red.
}
	\label{fig:long}
	\label{fig:onecol}
\end{figure}

\section{Introduction}
\label{sec:intro}

Weakly supervised object localization (WSOL) refers to a class of problems that aim to localize objects in an image using image-level annotations as weak supervision. Typically used image-level annotations are classification labels \cite{zhou2016learning}, image captions and action annotations \cite{selvaraju2017grad}, object size estimates \cite{shi2016weakly} and context-aware guidance \cite{kantorov2016contextlocnet}. In this paper, we use image classification labels as weak supervision to localize objects in images. Object location annotations are never explicitly used during training. 

We detail below common challenges observed in WSOL with classification labels. Figure \ref{fig:onecol} showcases common WSOL challenges and output of ViTOL for three query images.
(a) \textit{Highlighting Only Discriminative Regions}:
Fully supervised image classification does not accurately localize objects in an image as observed in the activation maps (AM) overlaid on Figure \ref{fig:onecol} $(column 1, row 2)$. We observe that the entire dog is not highlighted in the AM. In contrast, only the regions which the model considers as most-discriminative for classification are highlighted. This negatively affects the model's ability to localize the complete object of interest.
(b) \textit{Class Agnostic Activation Map}:
Figure \ref{fig:onecol} $(column 2, row 2)$ demonstrates that objects belonging to multiple classes are highlighted in the same AM. This limits the model's flexibility in localizing objects belonging to a specific class.
(c) \textit{Background Noise}: 
Accurately localizing an object of interest in a cluttered environment with many objects in the scene is challenging and results in a noisy AM as observed in Figure \ref{fig:onecol} $(column 3, row 2)$.

To alleviate the above challenges, we leverage vision transformer architectures \cite{dosovitskiy2020image} to localize objects in an image. Post-hoc work \cite{abnar2020quantifying, chefer2021transformer} has interpreted and visualized the capabilities of self-attention layers in transformers to understand objects of interest in a given image. In our work, we introduce below \textit{simple} modifications to the vision transformer. These modifications together with the vision transformer backbone enable a performance improvement of $6.57\%$ (MaxBox Accuracy-V2) and $5.24\%$ (Top-1 Localization) on ImageNet for WSOL over the current state-of-the-art. 

\textbf{patch-based Attention Dropout Layer (p-ADL)}: 
Attention dropout mechanism \cite{choe2019attention} is applied to a sequence of patch embeddings in ViTOL using our proposed p-ADL layer. Two components of the p-ADL layer are patch importance map and patch drop mask. During the course of training, these components are utilized to highlight informative and drop discriminative patches respectively to balance the classification v/s localization performance of the model. For a more detailed explanation, refer to Section \ref{sec:arch}.


\textbf{Grad Attention Rollout (GAR)}: 
Attention rollout \cite{abnar2020quantifying} generates an attention map by recursively multiplying the attention weight matrices across layers. We introduce a weighted attention rollout mechanism using the gradients computed at each attention map. We refer to this mechanism as Grad Attention Rollout. This post-hoc method in combination with the p-ADL enables the model to quantify the positive and negative contributions in the attention map for each class. This guides the model to generate class-dependent attention maps and a negative clamping operation in GAR further suppresses the effect of background noise in the attention map.
We observe that this method is effective in improving upon existing state-of-the-art for WSOL (refer Section \ref{sec:exps}). In Section \ref{sec:exps}, we also showcase results for an alternative post-hoc approach called Layer Relevance Propagation \cite{chefer2021transformer}.

\begin{figure*}
	\begin{center}
		\includegraphics[width=\linewidth, trim = 0 1.5cm 3cm 0, clip]{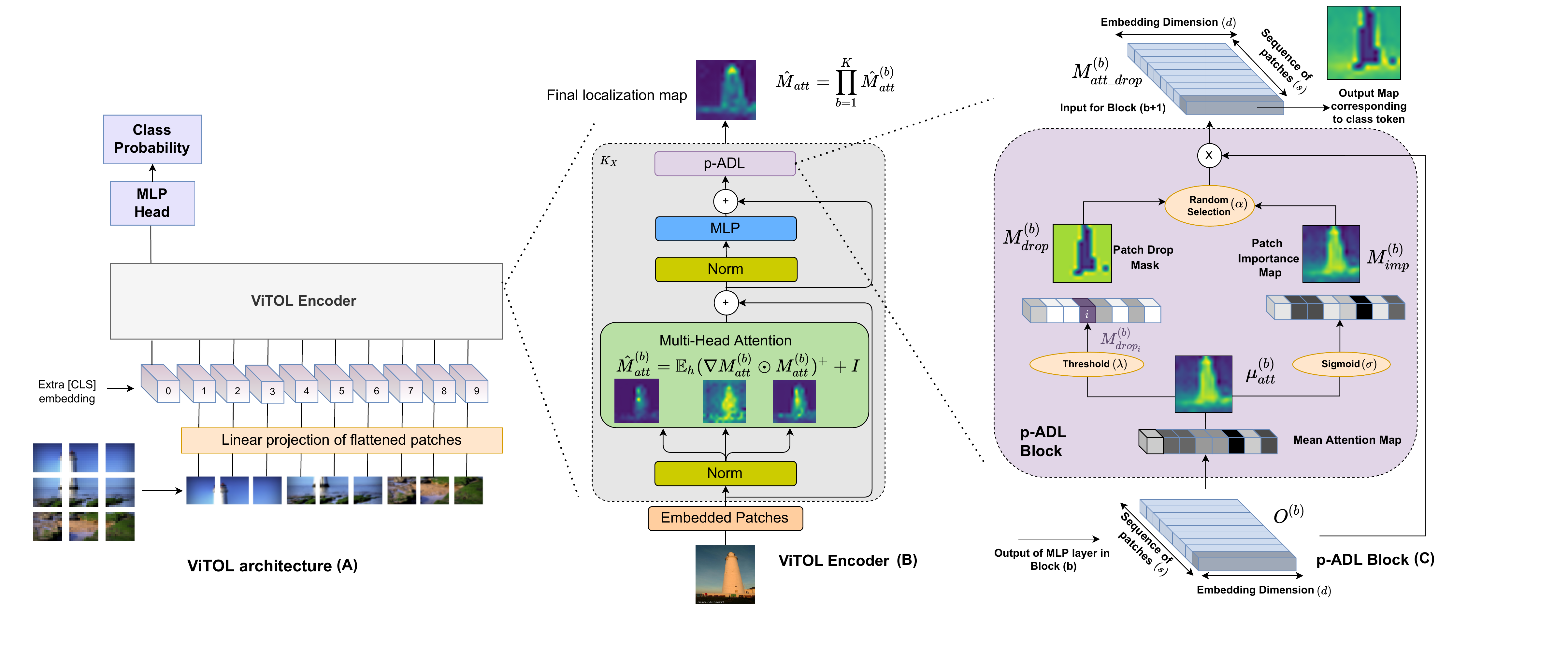}
	\end{center}
	\caption{Model overview: 
    (A) ViTOL architecture
	(B) ViTOL Encoder which consists of an additional p-ADL block. 
    (C) p-ADL Block and its components. 
    We visualize attention maps in B, p-ADL components in C and also highlight critical symbols and GAR equations. 
	}
	\label{fig:ViTOL}
	\vspace{-1.5em}
\end{figure*}

\textbf{Contributions:} In summary, we introduce three simple yet effective modifications to current WSOL methods that enable us to achieve state-of-the-art performance on object localization benchmarks. 1) we adopt a self-attention mechanism using vision transformer in the WSOL task, 2) introduce the p-ADL layer (Figure \ref{fig:ViTOL} (C)) to improve the localization ability of the vision transformer model significantly, and 3) leverage grad attention rollout to generate class-dependent activation maps.

We conduct extensive ablation experiments, qualitative and quantitative comparisons against state-of-the-art approaches on the object localization and classification performance of our model. Our best model has a MaxBoxAccV2 accuracy of $70.47\%$ on \textit{ImageNet-1K} \cite{russakovsky2015imagenet} and $73.17\%$ on \textit{Caltech-UCSD Birds-200-2011 (CUB)} \cite{welinder2010caltech} which are $6.57\%$ and $6.67\%$ better than the current state-of-the-art. Qualitatively, we observe that our approach generates high-quality localization maps that cover the entire object and are fairly robust to clutter in the scene (Figure \ref{fig:sota_comparison}, \ref{fig:CUB}, \ref{fig:ImageNet}). 

\section{Related Work}


In a weakly supervised learning setup, 
a major challenge while training classification models is that the classification loss pushes the network to look at the most discriminative features in an image and fails to localize the entire object. To address this issue, various methods \cite{zhang2018adversarial, zhou2016learning, bae2020rethinking} have introduced architectural changes in CNNs. 
More recently, the transformer architecture was introduced by Vaswani \etal \cite{vaswani2017attention} for machine translation and is widely used since then in many computer vision tasks. Vision transformers (ViT) \cite{dosovitskiy2020image} divide the image into a sequence of image patches and uses an extra class token to learn image representation and perform classification. A more powerful and data efficient extension of transformers was introduced as DeiT \cite{touvron2021training, khan2021transformers}.
Naseer \etal \cite{naseer2021intriguing} shows that the transformers showcase properties of handling long-range dependencies using their inherent self-attention mechanism. They preserve global information consistently across all layers (Fig.7 in \cite{dosovitskiy2020image}). Moreover, they exhibit properties of robustness to occlusions and preserve object structure which makes them suitable for the WSOL task. 
Therefore, we utilize the DeiT-Small-Variant (DeiT-S) and DeiT-Base-Variant (DeiT-B) transformer architecture in our work and introduce simple architectural changes to improve localization capabilities. 

In WSOL, some pioneering work has been proposed to generate localization maps. 
CAM \cite{zhou2016learning} uses a global average pooling layer \cite{lin2013network} to fuse feature maps and generate a localization map. This has a limitation of focusing only on the small discriminative part of the object. Wonho Bae et al. \cite{bae2020rethinking} improved CAM by using thresholded average pooling, negative weight clamping and percentile based thresholding techniques. TS-CAM \cite{gao2021ts} introduces a visual transformer based variant of CAM. They explicitly couple semantic agnostic maps from the vision transformer with semantic-aware maps from CAM to effectively localize objects in the image. In contrast, we introduce a grad attention rollout mechanism (see Sec. \ref{sec: maps}) in addition to architectural changes to the vision transformer which implicitly enable semantic aware object localization.
HaS \cite{kumar2017hide}, CutMix\cite{yun2019cutmix} drop some random parts in the input image to enforce the network to learn the importance of object sub-parts. Despite hiding the image region randomly, it does not necessarily erase the most discriminative parts. In contrast, our approach drops the most-discriminative regions in the attention maps to enforce localization of the entire object (see Sec. \ref{sec:arch}). 
Some CNN based recent work include ACoL \cite{zhang2018adversarial}, MEIL \cite{mai2020erasing}, SPG \cite{zhang2018self}, GC-Net \cite{lu2020geometry} and I2C \cite{zhang2020inter}. We empirically compare the localization performance against each of the above approaches in Sec. \ref{sec:exps}.
 

\textbf{Attention Dropout Mechanism}: ADL \cite{choe2019attention} applies attention based dropout mechanism to feature maps in deeper layers in CNNs. In our approach, we introduce a patch-based counterpart to ADL called p-ADL (see Section \ref{sec:arch}) which acts as a regularizer and enhances the localization capability of the model. Unlike ADL which operates on feature maps, p-ADL operates on the patch embeddings including the class token embedding.
Owing to the global attention property across all the layers in a transformer, we apply p-ADL at every transformer encoder block.


\textbf{Class Dependent Attention:} Attention maps also need to be class-dependent to correctly identify the object of interest in an image. 
For CNNs, class dependent behavior was introduced using gradient-based methods such as Grad-CAM \cite{selvaraju2017grad}, Grad-CAM++ \cite{chattopadhay2018grad}.
Furthermore, attribution propagation methods \cite{bach2015pixel, binder2016layer} propagate relevance for the predicted class based on Deep Taylor Decomposition (DTD) framework \cite{montavon2017explaining}. Class-specific behavior is also introduced using Contrastive-LRP \cite{gu2018understanding} and Softmax-Gradient LRP \cite{hooker2019benchmark}. 

For transformers, Abnar et al. \cite{abnar2020quantifying} proposed attention rollout method, which recursively multiply softmax activations corresponding to the class token in each attention layer. Our proposed method uses the gradients with respect to target class to generate final localization maps. 
Chefer et al. \cite{chefer2021transformer} proposed an alternate method of assigning local relevance based on the DTD principle to generate class dependent attention maps.
We evaluate attention rollout, gradient-based rollout and layer relevance propagation methods empirically and qualitatively in our proposed approach.

\section{Method}
In this section, we discuss the details of the proposed method, ViTOL, for WSOL. ViTOL mainly introduces (a) architectural changes and (b) localization map generation methods. 
In \textit{architectural changes}, we discuss changes to vision transformer architecture by introducing a  patch-based Attention Dropout Layer (p-ADL) as shown in Figure \ref{fig:ViTOL}.  
The p-ADL layer acts as a regularizer and assists the network to focus on both discriminative as well as non-discriminative regions of an object. This results in coverage of the entire object region, thereby greatly enhancing the localization ability.  
In \textit{map generation methods}, we discuss grad attention rollout mechanism (GAR). We compare this against attention rollout (AR) mechanism \cite{abnar2020quantifying} and layer relevance propogation (LRP) \cite{chefer2021transformer} for generating class dependent attention maps.
We discuss below both these changes in detail and also explain the procedure for generating high-quality localization maps.

\subsection{Architectural Changes} \label{sec:arch}

\textbf{Vision Transformer}: 
We leverage self-attention mechanism of vision transformers and employ DeiT-B and DeiT-S variants to our task.
Self-attention module \cite{vaswani2017attention} in transformers is defined below.

\begin{equation}
	M^{(b)}_{att} = softmax(Q^{(b)} \cdot {K^{(b)}}^T)
\end{equation}
\begin{equation}
	Z^{(b)} = M_{att}^{(b)} \cdot V^{(b)}
\end{equation}

$M^{(b)}_{att} \in \mathbb{R}^{(h \times s \times s)}$ is the attention matrix for encoder block $b$.
$Z_{(b)} \in \mathbb{R}^{(h \times s \times d_h)}$ is the output of the self-attention layer in block $b$.
Here, $h, s, d_h, d$ denotes the number of heads in each attention block, number of input sequence of patches, dimension of each head and embedding dimension of the input respectively. Generally, $d=h \cdot d_h$ is used. 
Here, notations follow \cite{chefer2021transformer}.

\textbf{patch-based Attention Dropout Layer (p-ADL)}: 
The p-ADL block is illustrated in Figure \ref{fig:ViTOL} (C). Largely inspired by the success of the attention-based dropout layer \cite{choe2019attention} on CNN-based architectures, we propose the \textbf{\textit{p-ADL}} layer for transformers. This layer has two main components, patch importance map and the patch drop mask. Both these components operate on the mean attention map, which is computed by taking the mean over the embedding dimension. \footnote{ We use the class token embedding while calculating the mean. This empirically results in a better localization performance.}  The \textit{patch drop mask} is created by dropping the most activated mean patch embedding based on a \textit{drop threshold ($\lambda$)} parameter. The \textit{patch importance map} is calculated by normalizing the mean attention map using a sigmoid activation,
which denotes the importance of patches in an image. 
 
\begin{figure}[h]
	\begin{center}
		\includegraphics[width=\linewidth, trim = 1cm 0.5cm 0.5cm 0 , clip]{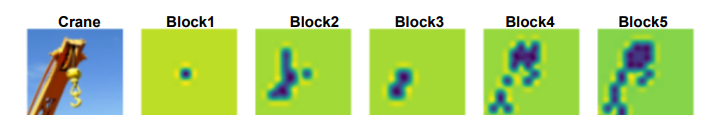}
	\end{center}
	\caption{\textbf{Patch Drop Mask progression}. Patch drop mask for set of encoder blocks are shown here for a sample crane image using our model with p-ADL layer. Dark color denotes dropped regions.
}
	\label{fig:drop_progression_crane}
	\vspace{-1em}
\end{figure}

Either the patch importance map or the patch drop mask is randomly chosen during training based on an \textit{embedding drop rate ($\alpha$)} parameter. Patch drop mask facilitates the network to improve its localization capability. This is because the most discriminative region of the attention map is dropped, forcing the model to focus on the remaining region of the desired object (as shown in Fig. \ref{fig:drop_progression_crane}).
Patch importance map aids the classification accuracy. This is because, over the course of training, the patch drop mask might drop informative regions of the object of interest. Patch importance map counters this behavior by ensuring necessary information about the object of interest is retained (see Figure \ref{fig:ViTOL} (C)). Some ablations regarding p-ADL layer are studied in Section \ref{subsection:ablation}. Note that this layer does not introduce any new trainable parameters. We formulate the following equations for the p-ADL layer.

\begin{equation}
	\mu_{att}^{(b)} = \mathbb{E}_d(O^{(b)}) 
\end{equation}
\begin{equation}
	M^{(b)}_{imp} = \sigma(\mu_{att}^{(b)})
\end{equation}
\begin{equation}
	{M^{(b)}_{drop}}_i = 
	\begin{cases} 
		0 & {\mu_{att}^{(b)}}_{i}\geq (\lambda \cdot max(\mu_{att}^{(b)})) \\
		1 & {\mu_{att}^{(b)}}_{i} < (\lambda \cdot max(\mu_{att}^{(b)}))\\
	\end{cases}
\end{equation}
\begin{equation}
	M^{(b)}_{att\_drop} = 
	\begin{cases} 
		O^{(b)} \cdot M^{(b)}_{drop} &  p_{random}<\alpha \\
		O^{(b)} \cdot M^{(b)}_{imp} &  p_{random}\geq\alpha \\
	\end{cases}
\end{equation}

where $O^{(b)} \in \mathbb{R}^{(s \times d)}$ denotes the output patch embedding matrix at encoder block $b$, $\mathbb{E}_d$ denotes the mean along embedding dimension, $\mu_{att}^{(b)}, M^{(b)}_{imp}, M^{(b)}_{drop} \in \mathbb{R}^{(s)}$ denotes the mean patch embedding, patch importance map, patch drop mask for encoder block $b$, ${M^{(b)}_{drop}}_i$ denotes the $i^{th}$ index of $M^{(b)}_{drop}$, $s$ is the number of patches in input sequence,  $M^{(b)}_{att\_drop} \in \mathbb{R}^{(s \times d)}$ is the output of the  \textit{p-ADL} layer for encoder block $b$. $\sigma$ denotes the sigmoid function. $p_{random} \in [0, 1]$ can be generated using any random number generator function.

\textbf{Patch drop mask intuition:} In Fig. \ref{fig:drop_progression_crane}, we visualize the output of patch drop mask corresponding to the class token, after every encoder block, for a query image.
Initially, a very small patch of crane was activated which is dropped in Block 1 by the patch drop mask. Dropping that information forces the model to explore other parts of crane, to help classify it correctly.  
This happens after each encoder block (Block $2$-$5$), thus driving the model to explore more object regions.  
However, dropping every informative patch progressively after each block might lead to embeddings which no longer have information about the object (Block $5$). This loss of object information may lead to object mis-classification. 
To avoid such circumstances, patch importance map is also chosen randomly which highlights the activated patches and retains object specific information. The visualization for patch importance and patch drop mask for an example image is observed in Figure \ref{fig:ViTOL} (C).

\begin{figure*}
    \begin{center} 
        \centering
        \captionsetup{type=figure}
        \includegraphics[width=0.8\textwidth, trim={0 0 0  0},clip]{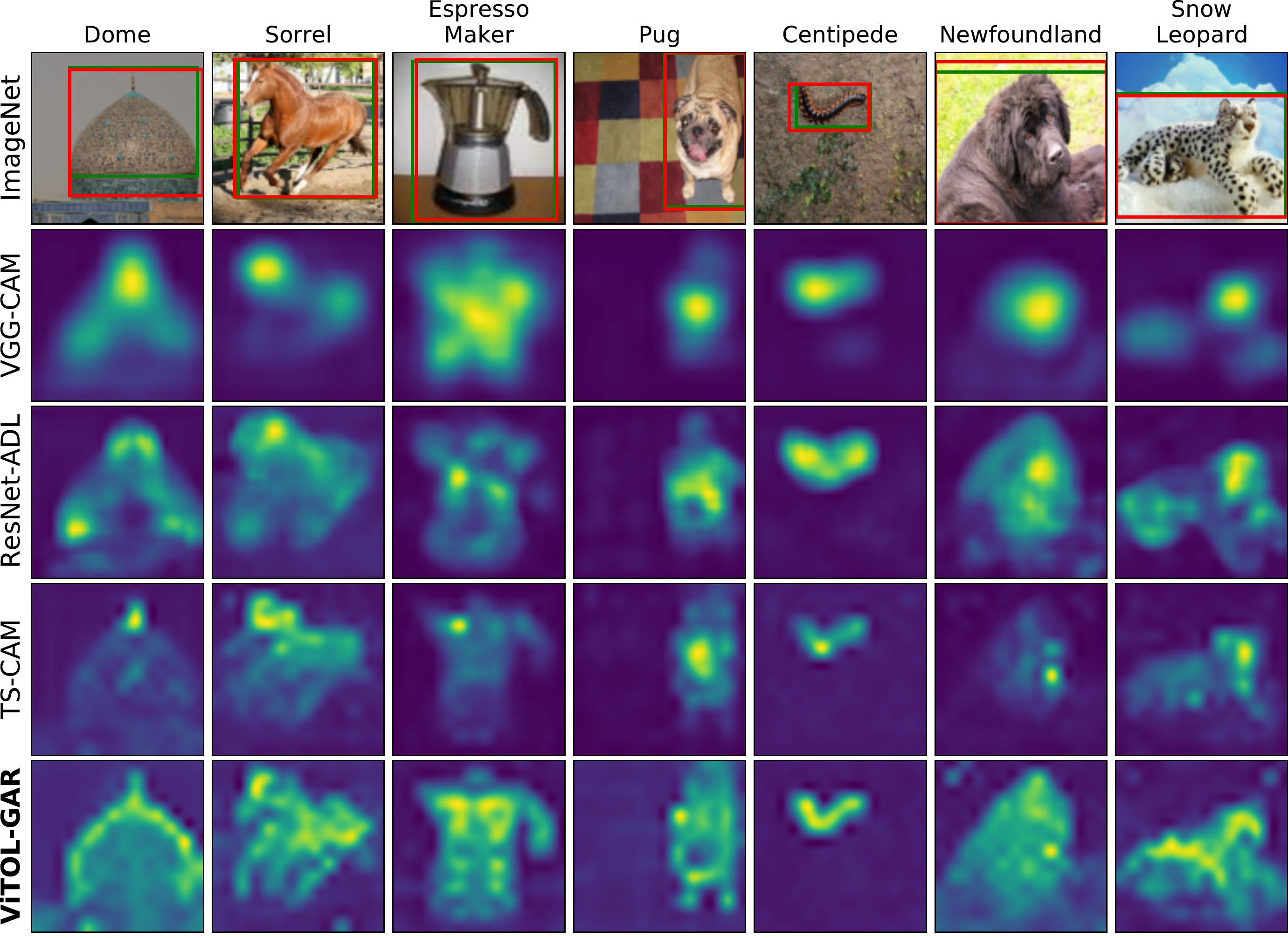}
        \captionof{figure}{\textbf{Qualitative comparison of activation maps with the state-of-the-art weakly supervised object localization models}. \\ Row-1: Query Image from ImageNet, Row-2: VGG16-CAM, Row-3: ResNet50-ADL, Row-4: TS-CAM based on DeiT-S backbone, Row-5: ViTOL based on Gradient Attention Rollout (GAR). Class label is displayed as a column label.}
        \label{fig:sota_comparison}
    \end{center}
    \vspace{0.4em}
\end{figure*}

\subsection{Generation of localization maps} \label{sec: maps}
\textbf{Grad Attention Rollout (GAR):} Attention rollout \cite{abnar2020quantifying} is an intuitive mechanism in which self-attention maps are averaged over each attention head in the encoder block and recursively multiplied over all transformer layers to produce a localization map. Rollout adds an Identity matrix $I$ to the attention matrix at each layer to account for residual connections \cite{he2016deep} in the network.  However, this method assumes that attentions are linearly combined. As stated by Chefer \etal \cite{chefer2021transformer}, this overlooks the fact that GELU \cite{hendrycks2016gaussian} activation is used in all intermediate layers. An ill-effect of this ignorance is that it fails to distinguish between positive and negative contributions to the final attention map. In our method, we compute a weighted attention rollout using gradient maps for each respective attention map and then use negative attention clamping to consider only positive values for each attention head. The GAR method is as given below. 

\begin{equation} \label{eq:GAR}
	\hat{M}^{(b)}_{att} = I + \mathbb{E}_h(\nabla{M}^{(b)}_{att} \odot M^{(b)}_{att})^+ 
\end{equation}
\begin{equation}{\label{rollout}}
	\hat{M}_{att} = \prod_{b=1}^{K} \hat{M}^{(b)}_{att}
\end{equation}

Here $\odot$ is the Hadamard product, and  $\mathbb{E}_h$ is the mean over head dimension, $\nabla{M}^{(b)}_{att} \in \mathbb{R}^{(h \times s \times s)}$ is the gradient matrix corresponding to each attention matrix ${M}^{(b)}_{att}$, K is the total number of encoder blocks. $\hat{M}_{att} \in \mathbb{R}^{(s \times s)} $ is the final rollout attention map. Superscript + in Eq. \ref{eq:GAR} refers to negative weight clamping operation.

\textbf{Layer Relevance Maps:} Chefer et al. \cite{chefer2021transformer} examined the problems in attention rollout and proposed a method that computes local relevance based on the deep Taylor decomposition principle.
This provides class-specific visualizations. For more details we refer the reader to \cite{chefer2021transformer}. The final attention map is computed as follows. 

\begin{equation} \label{eq:LRP}
	\Tilde{M}^{(b)}_{att} = I + \mathbb{E}_h(\nabla{M}^{(b)}_{att} \odot R^{(n_b)})^+
\end{equation}
\begin{equation}{\label{LRP}}
	\Tilde{M}_{att} = \prod_{b=1}^{K} \Tilde{M}^{(b)}_{att}
\end{equation}

where $\odot$ is the Hadamard product, $n_b$ denotes the softmax layer in encoder block $b$. $ R^{(n_b)}$ denotes the relevance for $n_b$ layer, $\Tilde{M}_{att} \in \mathbb{R}^{(s \times s)} $ is the final attention map.

In our approach, we evaluate both the methods defined above to get a self-attention map using Eq. \eqref{rollout}, \eqref{LRP}. This final matrix $M_{att}$ of size $(s \times s)$ represents how each patch in an image interacts with other patches. We only consider the first row of size $(1 \times s)$ corresponding to the $[CLS]$ token to compute the final attention map.  We skip the first value in this row to ignore the $[CLS]$ token's attention with itself and reshape the extracted attention embedding of size $(s-1)$ to size ${(\sqrt{s-1} \times \sqrt{s-1})}$ which denotes the final localization map for our method. 

\begin{table*}
	\begin{center}
		\begin{tabular}{|c|c|c|c|c|c|}
			\hline
			{} & \multicolumn{2}{|c|}{ImageNet} & \multicolumn{2}{|c|}{CUB}\\
			
			\hline
			Method & Architecture & MaxBoxAccV2 & Architecture & MaxBoxAccV2 \\
			\hline
			CAM & ResNet & 63.7 &  VGG & 63.7  \\
			\hline 
			HaS & Inception & 63.7 & ResNet & 64.7 \\
			\hline 
			ACoL & Inception & 63.7 & ResNet & 66.5 \\
			\hline 
			SPG & Inception & 63.3 & ResNet & 60.4  \\
			\hline 
			ADL & ResNet & 63.7 &  VGG & 66.3  \\
			\hline 
			CutMix & Inception & 63.9 &  ResNet & 62.8  \\
			\hline
			\textbf{Ours-GAR} & \textbf{DeiT-B+p-ADL} & \textbf{69.17}  & \textbf{DeiT-B+p-ADL} & \textbf{72.42} \\
			\hline
			\textbf{Ours-LRP} & \textbf{DeiT-B+p-ADL} & \textbf{70.47}  & \textbf{DeiT-B+p-ADL} &  \textbf{73.17} \\			
			
			\hline
		\end{tabular}
	\end{center}
	\caption{\textbf{Comparison with the state-of-the-art methods (MaxBoxAccV2)}. Scores for MaxBoxAccV2 for the baseline methods are borrowed from \cite{choe2020evaluation} on test split for both \textit{ImageNet} and \textit{CUB} datasets. 
}
	\label{table:SOTA_maxbox}
\end{table*}

\begin{table*}
	\begin{center}
		\begin{tabular}{|c|c|c|c|c|}
			\hline
			Method & Architecture & Top1-Loc & GT-Known & Top1-CLS\\
			\hline
			CAM & InceptionV3 & 46.3 & 62.7 & 70.6\\
			\hline
			ACoL & VGG16 & 45.8 & 63.0 & 64.5 \\
			\hline
			SPG & InceptionV3 & 48.6 & 64.7 & 71.1 \\
			\hline
			ADL & InceptionV3 & 48.7 & - & 61.2 \\
			\hline
			CutMix & VGG16 & 43.5 & - & 66.4 \\
			\hline
			MEIL \cite{mai2020erasing} & InceptionV3 & 49.5 & - & - \\
			\hline
			GC-Net \cite{lu2020geometry} & InceptionV3 & 49.1 & - & -\\
			\hline
			CAAM+SSAM \cite{babar2021look} & ResNet50 & 52.36 & 67.89 & -\\
			\hline
			SPA \cite{pan2021unveiling} & InceptionV3 & 52.7 & 68.33 & - \\
			\hline
			I$^2$C \cite{zhang2020inter} & InceptionV3 & 53.1 & 68.5 & -\\
			\hline
			TS-CAM \cite{gao2021ts} & DeiT-S & 53.4 & 67.6 & 74.3\\
			\hline
			\textbf{Ours-GAR} & \textbf{DeiT-S+p-ADL} & \textbf{54.74} & \textbf{71.86} & 71.84\\
			\hline
			\textbf{Ours-LRP} & \textbf{DeiT-S+p-ADL} & \textbf{53.62} & \textbf{70.48} & 71.84\\
			\hline
			\textbf{Ours-GAR} & \textbf{DeiT-B+p-ADL} & \textbf{57.62} & \textbf{71.32} & \textbf{77.08}\\
			\hline
			\textbf{Ours-LRP} & \textbf{DeiT-B+p-ADL} & \textbf{58.64} & \textbf{72.51} & \textbf{77.08}\\
			\hline

		\end{tabular}
	\end{center}
	\caption{\textbf{Comparison with the state-of-the-art methods (Top1-Loc and GT-Known):} Top1-Loc and GT-Known metrics are compared with other WSOL methods on ImageNet validation data. }
	\label{table:SOTA_top1}
\end{table*}

\begin{table*}[]
	\begin{center}
	\begin{tabular}{|c|c|cccc|cccc|}
		\hline
		\multicolumn{1}{|l|}{} &                       & \multicolumn{4}{c|}{ImageNet}                                                                                                   & \multicolumn{4}{c|}{CUB}                                                                                                         \\ \hline
		Architecture           & Method & \multicolumn{1}{c|}{MaxboxAccV2}    & \multicolumn{1}{c|}{IOU30}         & \multicolumn{1}{c|}{IOU50}          & IOU70          & \multicolumn{1}{c|}{MaxboxAccV2}    & \multicolumn{1}{c|}{IOU30}          & \multicolumn{1}{c|}{IOU50}          & IOU70          \\ \hline
		DeiT-B                 & AR               & \multicolumn{1}{c|}{45.94}          & \multicolumn{1}{c|}{70.41}         & \multicolumn{1}{c|}{44.39}          & 23.03          & \multicolumn{1}{c|}{49.75}          & \multicolumn{1}{c|}{83.27}          & \multicolumn{1}{c|}{49.37}          & 16.6           \\ \hline
		DeiT-B+p-ADL         & AR               & \multicolumn{1}{c|}{46.51}          & \multicolumn{1}{c|}{70.71}         & \multicolumn{1}{c|}{45.33}         & 23.47         & \multicolumn{1}{c|}{57.42}          & \multicolumn{1}{c|}{88.45}          & \multicolumn{1}{c|}{60.87}          & 22.93          \\ \hline
		DeiT-B                 & GAR                   & \multicolumn{1}{c|}{62.82}          & \multicolumn{1}{c|}{83.58}         & \multicolumn{1}{c|}{64.89}          & 40.00          & \multicolumn{1}{c|}{70.79}          & \multicolumn{1}{c|}{96.15}          & \multicolumn{1}{c|}{78.2}           & 38.03          \\ \hline
		DeiT-B+p-ADL         & GAR                   & \multicolumn{1}{c|}{\textbf{69.17}} & \multicolumn{1}{c|}{\textbf{86.95}} & \multicolumn{1}{c|}{\textbf{71.32}} & \textbf{49.25} & \multicolumn{1}{c|}{\textbf{72.42}} & \multicolumn{1}{c|}{\textbf{96.68}} & \multicolumn{1}{c|}{\textbf{80.89}} & \textbf{39.69} \\ \hline
	\end{tabular}
	\end{center}
	\caption{\textbf{Effect of p-ADL and Comparison of Attention Rollout v/s Grad Attention Rollout }}
	\label{table:rollout_vs_grad}
	\vspace{-1em}
\end{table*}

\begin{table}[b]
	\begin{center}
		\begin{tabular}{|c|c|c|c|c|}
			\hline
			Architecture & MMp & MpM & ImageNet & CUB \\
			\hline
			DeiT-B+ p-ADL & {} & \checkmark & 70.41 & 71.99 \\
			\hline			
			DeiT-B+ p-ADL& \checkmark & {} & 70.47 & 73.17 \\
			\hline
		\end{tabular}
	\end{center}
	\caption{\textbf{Changing p-ADL position inside encoder blocks:} MMp [MSA $\ll$ MLP $\ll$ p-ADL] denotes that p-ADL is placed after MLP inside each encoder block and MpM [MSA $\ll$ p-ADL $\ll$ MLP] denotes that p-ADL is inserted in between MLP and MSA layers. MaxBoxAccv2 is compared on both ImageNet and CUB.}
	\label{table:p-ADL_position}
\end{table}

\section{Experiments} \label{sec:exps}
\subsection{Dataset} We evaluate ViTOL on two benchmark datasets in WSOL namely ImageNet-1K \cite{russakovsky2015imagenet} and Caltech-UCSD Birds-200-2011 (CUB) \cite{welinder2010caltech}. We use the same data splits as defined in \cite{choe2020evaluation}. Imagenet-1K dataset has 1000 classes with approximately 1.3M images for training and 50k images for testing. It is one of the most challenging datasets because of its large inter-class variation.
Previous WSOL methods \cite{zhou2016learning, kumar2017hide, zhang2018adversarial, zhang2018self, choe2019attention, yun2019cutmix} have seen relatively small improvements (MaxBoxAccV2) for WSOL as observed in Table \ref{table:SOTA_maxbox}. CUB dataset is comparatively small with 200 classes consisting of 5994 train and 5794 test images.


\subsection{Evaluation Metric}
We generate a binary mask from localization map and use it to generate a prediction bounding box around the foreground. If the Intersection over Union (IOU) of the predicted box and ground truth box is greater than a threshold $\delta$ then it is classified as a correct prediction. We use 4 metrics for evaluating all our methods: 1) \textit{MaxBoxAccV2 \cite{choe2020evaluation}:}  average of the localization accuracy across IOU thresholds $\delta \in \{0.3, 0.5, 0.7\}$ to show the fineness of localization. 2) \textit{GT-known Loc (IOU 50):} use the GT label during inference to measure performance at fixed IOU threshold $\delta=0.5$. 3) \textit{Top-1-Loc}: This metric provides a positive value when classification is correct and IOU is greater than $50\%$. 4) \textit{Top1-CLS:} Top1 classification performance.

\textbf{Key Metric}: \textit{MaxBoxAccV2} metric measures the ability of the model to localize the objects of interest, \textit{which we consider as the key criterion to compare different WSOL approaches}. Choe \etal \cite{choe2020evaluation} argue that the \textit{goal of WSOL is to localize objects and not classify images and therefore they advocate the use of MaxBoxAccV2} as the desired performance metric. We adopt this recently fixed version of the localization metric. Moreover, only an increase in the classification accuracy without any observed improvement in the localization performance will increase Top1-localization accuracy value, however, \textit{this does not translate to the performance improvements on the localization task}. 

\subsection{Implementation Details}
We use DeiT-B and DeiT-S architectures for our task. We use $s=197$, $K=12$, patch size of $16$x$16$ and image resolution of $224$x$224$ in all our experiments. DeiT-S and DeiT-B uses embedding dimension of $384$ and $768$, and attention heads $6$ and $12$ respectively in each ViTOL encoder block. In p-ADL, we use a drop threshold ($\lambda$) of $0.9$  and embedding drop-rate ($\alpha$) as $0.75$ across all experiments.

\textbf{Training and Testing details}: On ImageNet-1K, for baseline models, we use the DeiT-B and DeiT-S ImageNet pre-trained weights and evaluate on all the methods, namely, AR, GAR  and LRP as stated in Table \ref{table:SOTA_maxbox}, \ref{table:SOTA_top1} and \ref{table:rollout_vs_grad}. In experiments with ViTOL, on ImageNet-1K, we train the model with the p-ADL layer for 10 epochs using a learning rate of 0.0001, a weight decay (wd) of 0, and a learning rate (lr) decay of 0.1 after every 3 epochs. For experiments on CUB dataset, to create a new baseline model, we take the ImageNet pre-trained DeiT-B model but initialize a new classifier layer with 200 class output. We trained the model for 50 epochs with a lr of 0.0001, wd of 0, and lr decay of 0.1 every 10 epochs. In experiments with ViTOL, we add the p-ADL layer and continue training this model. Here, we use lr of 0.00001, wd of 0, and lr decay of 0.1 after every 5 epochs. Following \cite{choe2020evaluation}, we use last saved checkpoint model for all the evaluations in Table \ref{table:SOTA_maxbox} and \ref{table:SOTA_top1}. During inference, p-ADL layer is not used.

\subsection{Quantitative Results}

\subsubsection{Comparison to the State-of-the-Art}
Table \ref{table:SOTA_maxbox} and Table \ref{table:SOTA_top1} compare our proposed approach ViTOL against the state-of-the-art (SOTA) approaches for WSOL. In Table \ref{table:SOTA_maxbox}, we use the  MaxBoxAccV2 metric \cite{choe2020evaluation} to quantify the localization performance on ImageNet and CUB datasets. Here, we compare against all SOTA approaches which report the aforementioned metrics. The \textit{Architecture} column mentions the network architecture across all approaches based on the best performing backbones on each dataset. We observe that our approaches with a DeiT-B backbone with p-ADL  + (a) GAR and (b) LRP significantly outperform the other WSOL approaches. Using the MaxBoxAccV2 metric we observe gains by a margin of $6.57\%$ and $6.67\%$, on ImageNet and CUB datasets respectively. 

In Table \ref{table:SOTA_top1}, we use the Top-1 localization accuracy metric to compare ViTOL against the SOTA approaches for WSOL on the ImageNet dataset. We outperform the SOTA approaches by achieving a Top-1 localization accuracy of $58.64\%$, a GT-Known value of $72.51\%$ and Top-1 classification (CLS) accuracy of $77.08\%$.

\begin{figure}[h]
	\begin{center}
		\includegraphics[width=\linewidth, trim = 4.5cm 2cm 4cm 1cm, clip]{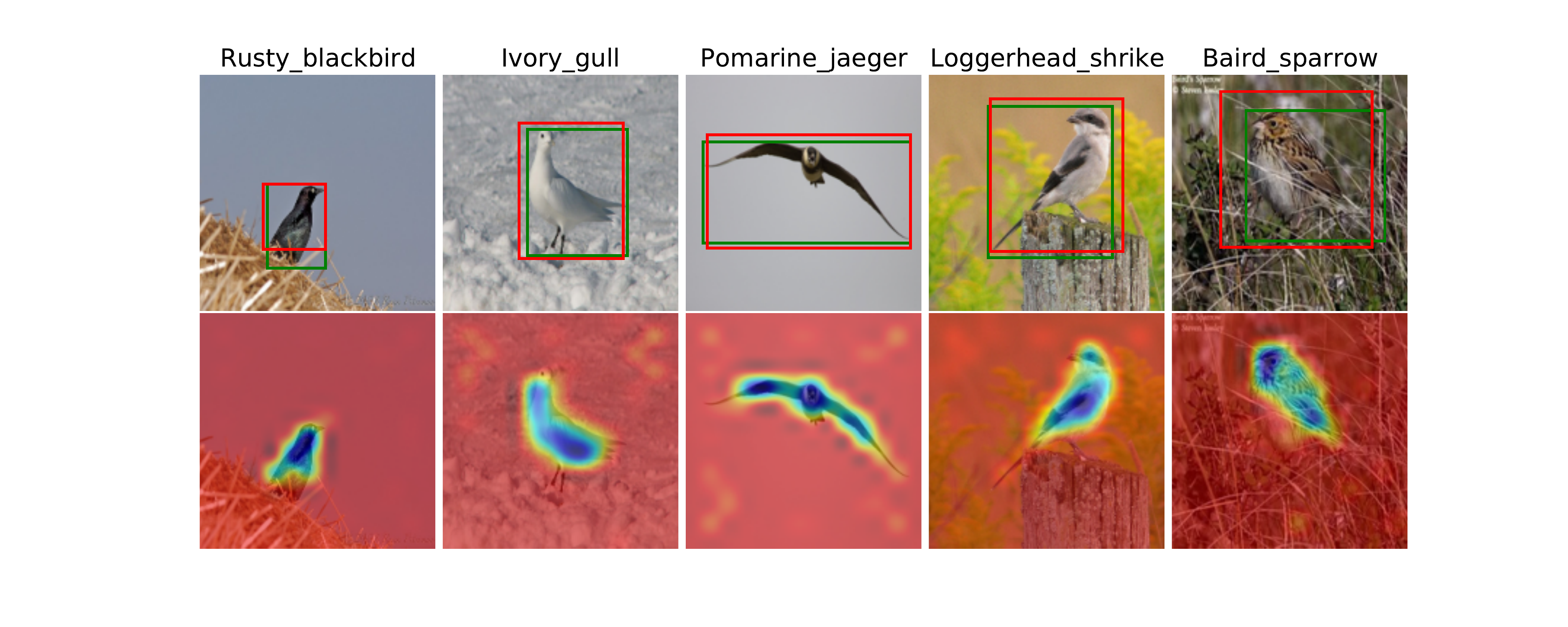}
	\end{center}
	\caption{\textbf{Qualitative results on CUB}. GT bounding box is shown in green and predicted bounding box is shown in red.  
}
	\label{fig:CUB}
    \vspace{-1em}
\end{figure}

\subsubsection{Ablation Study} \label{subsection:ablation}
\textbf{Impact of p-ADL}:
In Table \ref{table:rollout_vs_grad}, we compare the performance of DeiT-B with and without p-ADL using AR and GAR map generation mechanisms. We observe that, 
for the AR map-generation method, the DeiT-B+p-ADL achieves a perfomance gain of $0.57\%$ and $7.67\%$ on ImageNet and CUB datasets respectively against DeiT-B. Similarly, for the GAR method, DeiT-B+p-ADL achieves a gain of $6.35\%$ and $1.63\%$ respectively against DeiT-B. These performance gains can be attributed to the ability of the p-ADL layer to highlight discriminative and non-discriminative regions of the object of interest.

\textbf{Position of p-ADL}:
We experiment with the position of the p-ADL layer inside each encoder block to understand its impact on the localization performance. We observe from Table \ref{table:p-ADL_position} that we get minor performance improvements if we place the p-ADL after the MLP layer as shown in Fig \ref{fig:ViTOL}. This minor difference can be intuitively attributed to the MLP layer prioritizing classification performance over the localization in the case of p-ADL placed before MLP.

\textbf{Improvements with GAR}:
The usage of gradients (equation \ref{eq:GAR}) provides additional information about the semantics of the class of the object. This change introduces class dependent behavior for the attention maps. We observe a huge performance gain attributed to this behavior in Table \ref{table:rollout_vs_grad}. A gain of $22.66\%$ and $15\%$ is observed for ImageNet and CUB datasets respectively for the DeiT+p-ADL model.

\textbf{Other ablations:}
Some other ablation studies include i) comparison of GAR and LRP across different transformer backbones, ii) detailed study of patch drop masks for the p-ADL layer and iii) effect of changing embedding drop rate and drop threshold hyper-parameters for p-ADL layer. We refer the reader to supplementary for these ablations.

\begin{figure}[t]
	\begin{center}
		\includegraphics[width=\linewidth, trim = 3cm 3cm 3.5cm 1cm, clip]{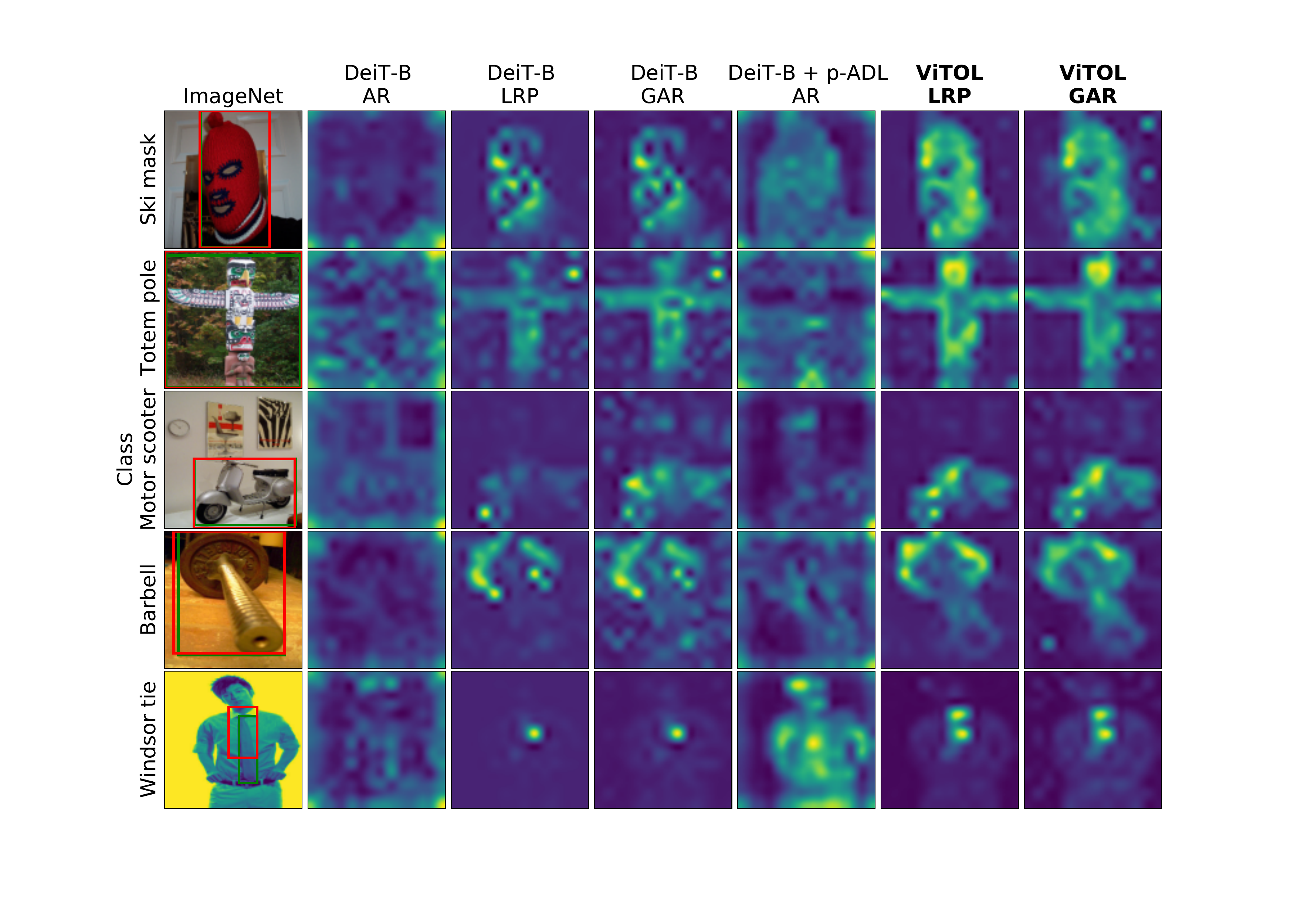}
	\end{center}
	\caption{\textbf{Qualitative ablation for map generation methods}. Comparison of final localization map for baseline DeiT-B (Column 2,3,4) and DeiT-B $+$ p-ADL layer (Column 5, 6, 7) is shown here for map generation methods AR, GAR and LRP. GT bounding box is shown in green. Predicted bounding box (red) is calculated using the ViTOL-GAR. 
}
\vspace{-1.5em}
	\label{fig:ImageNet}
\end{figure}

\subsection{Qualitative Results}
In this section, we qualitatively compare the performance of ViTOL in Fig. \ref{fig:sota_comparison}. The attention maps of our model are visualized for CUB in Fig. \ref{fig:CUB} and ImageNet in Fig. \ref{fig:ImageNet}.

\textbf{Comparison with SOTA}:  In Figure \ref{fig:sota_comparison}, we highlight the localization maps for ViTOL and compare against state-of-the-art WSOL methods, VGG-CAM \cite{zhou2016learning}, ResNet50-ADL \cite{choe2019attention}, TS-CAM \cite{gao2021ts}, for a few randomly sampled images from ImageNet. Our method, ViTOL, shows superior localization performance in highlighting the entire object region across all the sample images. Moreover, our localization maps reconstruct the structure of the sorrel and centipede classes more accurately as compared with other methods.


\textbf{Qualitative Ablation on ImageNet}: In Figure \ref{fig:ImageNet}, we generate attention maps for five random example images from the ImageNet validation split. For each of the example images, we visualize the baseline DeiT-B with AR, GAR and LRP in the second, third and fourth columns, and, DeiT-B + p-ADL with AR, ViTOL with LRP and ViTOL with GAR attention maps in the final three columns.

We observe that attention maps generated for ViTOL with GAR/LRP show dependency with the class, are noise-free and cover the complete object of interest. Our method is superior in localizing the objects in the image as compared to the DeiT-B baseline models.  
In Figure \ref{fig:ImageNet}, row five, we observe that the attention maps using DeiT-B  highlight the whole person. However, the object of interest is the tie worn by the person. The class conditional behavior of ViTOL enables the models to approximately localize the \textit{tie} in the image. Similarly in row three, our method highlights the desired class in an image containing multiple objects. 
The first four rows showcase that the entire object of interest is highlighted in the ViTOL attention maps. In the second row we observe that the negative contributions from the noisy background observed in the attention maps of DeiT-B + p-ADL + AR are alleviated in ViTOL-GAR/LRP.

\textbf{Visualization on CUB}: In Figure \ref{fig:CUB}, first row, we showcase five random example images from the CUB dataset. In the second row, we overlay the attention maps obtained from ViTOL-LRP for these images. In the first column, half the bird is occluded and our model only captures the region of bird that is visible. In the second column, our method successfully localizes a bird camouflaged into its background. In the third column our method captures the bird's narrow wings in its attention map. In the fourth column, the tail of the bird looks similar to the stem of the plants in the background. Our method succeeds in highlighting the entire bird along with its thin tail. In last column, our method localizes the visible portion of the bird which is resting in a noisy background.
\section{Conclusion}


In this work, we explore the problem of WSOL. We discuss in detail the common challenges that image classification models encounter when localizing objects of interest. To alleviate these challenges, we introduced our proposed method \textbf{ViTOL}. We discuss simple yet effective changes to a baseline transformer architecture and introduced a new map generation method for WSOL. We employed patch-based attention dropout layer ({p-ADL}) which aided in increasing the coverage of the localization map. We also proposed a post-hoc map generation method which enforced class-dependent attention map generation for WSOL. Finally, we showcased state-of-the-art performance across different metrics, datasets and methods for object localization. Our experimental results establish the significance of using ViTOL in object localization. The clear resemblance of the actual shape of the object in the attention map visualization shows great potential to leverage it for other tasks such as weakly-supervised semantic segmentation, temporal action localization, etc. In order to apply the proposed p-ADL and GAR map generation methods for CNN based architectures, the p-ADL layer and rollout mechanism can potentially be applied to those layers that have dense features and high receptive fields. In future work,  we plan to explore variants of global attention in transformers to capture fine-grained details as opposed to the patch-based attention which we currently employ. 

\textbf{Acknowledgements:} This work was funded by Mercedes-Benz Research and Development, India. Special thanks to Brijesh Pillai for the research support and Sujay Babruwad for the assistance in our work.

{\small
\bibliographystyle{ieee_fullname}
\bibliography{egbib}

\begin{thebibliography}{10}\itemsep=-1pt

\bibitem{abnar2020quantifying}
Samira Abnar and Willem~H Zuidema.
\newblock Quantifying attention flow in transformers.
\newblock In {\em ACL}, 2020.

\bibitem{babar2021look}
Sadbhavana Babar and Sukhendu Das.
\newblock Where to look?: Mining complementary image regions for weakly
  supervised object localization.
\newblock In {\em Proceedings of the IEEE/CVF Winter Conference on Applications
  of Computer Vision}, pages 1010--1019, 2021.

\bibitem{bach2015pixel}
Sebastian Bach, Alexander Binder, Gr{\'e}goire Montavon, Frederick Klauschen,
  Klaus-Robert M{\"u}ller, and Wojciech Samek.
\newblock On pixel-wise explanations for non-linear classifier decisions by
  layer-wise relevance propagation.
\newblock {\em PloS one}, 10(7):e0130140, 2015.

\bibitem{bae2020rethinking}
Wonho Bae, Junhyug Noh, and Gunhee Kim.
\newblock Rethinking class activation mapping for weakly supervised object
  localization.
\newblock In {\em European Conference on Computer Vision}, pages 618--634.
  Springer, 2020.

\bibitem{binder2016layer}
Alexander Binder, Gr{\'e}goire Montavon, Sebastian Lapuschkin, Klaus-Robert
  M{\"u}ller, and Wojciech Samek.
\newblock Layer-wise relevance propagation for neural networks with local
  renormalization layers.
\newblock In {\em International Conference on Artificial Neural Networks},
  pages 63--71. Springer, 2016.

\bibitem{chattopadhay2018grad}
Aditya Chattopadhay, Anirban Sarkar, Prantik Howlader, and Vineeth~N
  Balasubramanian.
\newblock Grad-cam++: Generalized gradient-based visual explanations for deep
  convolutional networks.
\newblock In {\em 2018 IEEE Winter Conference on Applications of Computer
  Vision (WACV)}, pages 839--847. IEEE, 2018.

\bibitem{chefer2021transformer}
Hila Chefer, Shir Gur, and Lior Wolf.
\newblock Transformer interpretability beyond attention visualization.
\newblock In {\em Proceedings of the IEEE/CVF Conference on Computer Vision and
  Pattern Recognition}, pages 782--791, 2021.

\bibitem{choe2020evaluation}
Junsuk Choe, Seong~Joon Oh, Sanghyuk Chun, Zeynep Akata, and Hyunjung Shim.
\newblock Evaluation for weakly supervised object localization: Protocol,
  metrics, and datasets.
\newblock {\em arXiv preprint arXiv:2007.04178}, 2020.

\bibitem{choe2019attention}
Junsuk Choe and Hyunjung Shim.
\newblock Attention-based dropout layer for weakly supervised object
  localization.
\newblock In {\em Proceedings of the IEEE/CVF Conference on Computer Vision and
  Pattern Recognition}, pages 2219--2228, 2019.

\bibitem{dosovitskiy2020image}
Alexey Dosovitskiy, Lucas Beyer, Alexander Kolesnikov, Dirk Weissenborn,
  Xiaohua Zhai, Thomas Unterthiner, Mostafa Dehghani, Matthias Minderer, Georg
  Heigold, Sylvain Gelly, et~al.
\newblock An image is worth 16x16 words: Transformers for image recognition at
  scale.
\newblock In {\em International Conference on Learning Representations}, 2020.

\bibitem{gao2021ts}
Wei Gao, Fang Wan, Xingjia Pan, Zhiliang Peng, Qi Tian, Zhenjun Han, Bolei
  Zhou, and Qixiang Ye.
\newblock Ts-cam: Token semantic coupled attention map for weakly supervised
  object localization.
\newblock In {\em Proceedings of the IEEE/CVF International Conference on
  Computer Vision}, pages 2886--2895, 2021.

\bibitem{gu2018understanding}
Jindong Gu, Yinchong Yang, and Volker Tresp.
\newblock Understanding individual decisions of cnns via contrastive
  backpropagation.
\newblock In {\em Asian Conference on Computer Vision}, pages 119--134.
  Springer, 2018.

\bibitem{he2016deep}
Kaiming He, Xiangyu Zhang, Shaoqing Ren, and Jian Sun.
\newblock Deep residual learning for image recognition.
\newblock In {\em Proceedings of the IEEE conference on computer vision and
  pattern recognition}, pages 770--778, 2016.

\bibitem{hendrycks2016gaussian}
Dan Hendrycks and Kevin Gimpel.
\newblock Gaussian error linear units (gelus).
\newblock {\em arXiv preprint arXiv:1606.08415}, 2016.

\bibitem{hooker2019benchmark}
Sara Hooker, Dumitru Erhan, Pieter-Jan Kindermans, and Been Kim.
\newblock A benchmark for interpretability methods in deep neural networks.
\newblock {\em Advances in neural information processing systems}, 32, 2019.

\bibitem{kantorov2016contextlocnet}
Vadim Kantorov, Maxime Oquab, Minsu Cho, and Ivan Laptev.
\newblock Contextlocnet: Context-aware deep network models for weakly
  supervised localization.
\newblock In {\em European Conference on Computer Vision}, pages 350--365.
  Springer, 2016.

\bibitem{khan2021transformers}
Salman Khan, Muzammal Naseer, Munawar Hayat, Syed~Waqas Zamir, Fahad~Shahbaz
  Khan, and Mubarak Shah.
\newblock Transformers in vision: A survey.
\newblock {\em ACM Computing Surveys (CSUR)}, 2021.

\bibitem{kumar2017hide}
Krishna Kumar~Singh and Yong Jae~Lee.
\newblock Hide-and-seek: Forcing a network to be meticulous for
  weakly-supervised object and action localization.
\newblock In {\em Proceedings of the IEEE International Conference on Computer
  Vision}, pages 3524--3533, 2017.

\bibitem{lin2013network}
Min Lin, Qiang Chen, and Shuicheng Yan.
\newblock Network in network.
\newblock {\em arXiv preprint arXiv:1312.4400}, 2013.

\bibitem{lu2020geometry}
Weizeng Lu, Xi Jia, Weicheng Xie, Linlin Shen, Yicong Zhou, and Jinming Duan.
\newblock Geometry constrained weakly supervised object localization.
\newblock In {\em Computer Vision--ECCV 2020: 16th European Conference,
  Glasgow, UK, August 23--28, 2020, Proceedings, Part XXVI 16}, pages 481--496.
  Springer, 2020.

\bibitem{mai2020erasing}
Jinjie Mai, Meng Yang, and Wenfeng Luo.
\newblock Erasing integrated learning: A simple yet effective approach for
  weakly supervised object localization.
\newblock In {\em Proceedings of the IEEE/CVF Conference on Computer Vision and
  Pattern Recognition}, pages 8766--8775, 2020.

\bibitem{montavon2017explaining}
Gr{\'e}goire Montavon, Sebastian Lapuschkin, Alexander Binder, Wojciech Samek,
  and Klaus-Robert M{\"u}ller.
\newblock Explaining nonlinear classification decisions with deep taylor
  decomposition.
\newblock {\em Pattern Recognition}, 65:211--222, 2017.

\bibitem{naseer2021intriguing}
Muhammad~Muzammal Naseer, Kanchana Ranasinghe, Salman~H Khan, Munawar Hayat,
  Fahad Shahbaz~Khan, and Ming-Hsuan Yang.
\newblock Intriguing properties of vision transformers.
\newblock {\em Advances in Neural Information Processing Systems}, 34, 2021.

\bibitem{pan2021unveiling}
Xingjia Pan, Yingguo Gao, Zhiwen Lin, Fan Tang, Weiming Dong, Haolei Yuan,
  Feiyue Huang, and Changsheng Xu.
\newblock Unveiling the potential of structure preserving for weakly supervised
  object localization.
\newblock In {\em Proceedings of the IEEE/CVF Conference on Computer Vision and
  Pattern Recognition}, pages 11642--11651, 2021.

\bibitem{russakovsky2015imagenet}
Olga Russakovsky, Jia Deng, Hao Su, Jonathan Krause, Sanjeev Satheesh, Sean Ma,
  Zhiheng Huang, Andrej Karpathy, Aditya Khosla, Michael Bernstein, et~al.
\newblock Imagenet large scale visual recognition challenge.
\newblock {\em International journal of computer vision}, 115(3):211--252,
  2015.

\bibitem{selvaraju2017grad}
Ramprasaath~R Selvaraju, Michael Cogswell, Abhishek Das, Ramakrishna Vedantam,
  Devi Parikh, and Dhruv Batra.
\newblock Grad-cam: Visual explanations from deep networks via gradient-based
  localization.
\newblock In {\em Proceedings of the IEEE international conference on computer
  vision}, pages 618--626, 2017.

\bibitem{shi2016weakly}
Miaojing Shi and Vittorio Ferrari.
\newblock Weakly supervised object localization using size estimates.
\newblock In {\em European Conference on Computer Vision}, pages 105--121.
  Springer, 2016.

\bibitem{touvron2021training}
Hugo Touvron, Matthieu Cord, Matthijs Douze, Francisco Massa, Alexandre
  Sablayrolles, and Herv{\'e} J{\'e}gou.
\newblock Training data-efficient image transformers \& distillation through
  attention.
\newblock In {\em International Conference on Machine Learning}, pages
  10347--10357. PMLR, 2021.

\bibitem{vaswani2017attention}
Ashish Vaswani, Noam Shazeer, Niki Parmar, Jakob Uszkoreit, Llion Jones,
  Aidan~N Gomez, {\L}ukasz Kaiser, and Illia Polosukhin.
\newblock Attention is all you need.
\newblock {\em Advances in neural information processing systems}, 30, 2017.

\bibitem{welinder2010caltech}
Peter Welinder, Steve Branson, Takeshi Mita, Catherine Wah, Florian Schroff,
  Serge Belongie, and Pietro Perona.
\newblock Caltech-ucsd birds 200.
\newblock 2010.

\bibitem{yun2019cutmix}
Sangdoo Yun, Dongyoon Han, Seong~Joon Oh, Sanghyuk Chun, Junsuk Choe, and
  Youngjoon Yoo.
\newblock Cutmix: Regularization strategy to train strong classifiers with
  localizable features.
\newblock In {\em Proceedings of the IEEE/CVF International Conference on
  Computer Vision}, pages 6023--6032, 2019.

\bibitem{zhang2018adversarial}
Xiaolin Zhang, Yunchao Wei, Jiashi Feng, Yi Yang, and Thomas~S Huang.
\newblock Adversarial complementary learning for weakly supervised object
  localization.
\newblock In {\em Proceedings of the IEEE Conference on Computer Vision and
  Pattern Recognition}, pages 1325--1334, 2018.

\bibitem{zhang2018self}
Xiaolin Zhang, Yunchao Wei, Guoliang Kang, Yi Yang, and Thomas Huang.
\newblock Self-produced guidance for weakly-supervised object localization.
\newblock In {\em Proceedings of the European conference on computer vision
  (ECCV)}, pages 597--613, 2018.

\bibitem{zhang2020inter}
Xiaolin Zhang, Yunchao Wei, and Yi Yang.
\newblock Inter-image communication for weakly supervised localization.
\newblock 2020.

\bibitem{zhou2016learning}
Bolei Zhou, Aditya Khosla, Agata Lapedriza, Aude Oliva, and Antonio Torralba.
\newblock Learning deep features for discriminative localization.
\newblock In {\em Proceedings of the IEEE conference on computer vision and
  pattern recognition}, pages 2921--2929, 2016.

\end{thebibliography}
}

\end{document}


\title{Supplementary Material}

\maketitle



In this supplementary manuscript, we showcase (i) qualitative comparison of final localization maps between different weakly supervised object localization (WSOL) methods for a few query images, (ii) intuitive comparison of GAR and LRP across different transformer backbones, iii) comparison of the patch drop mask visualization across the transformer blocks with respect to the base transformer model DeiT-B, (iv) ablation of hyper-parameters chosen for the p-ADL layer.


\section{Qualitative Comparison with SOTA methods}

\begin{figure*}[t]
\begin{center}
\includegraphics[width=1\linewidth]{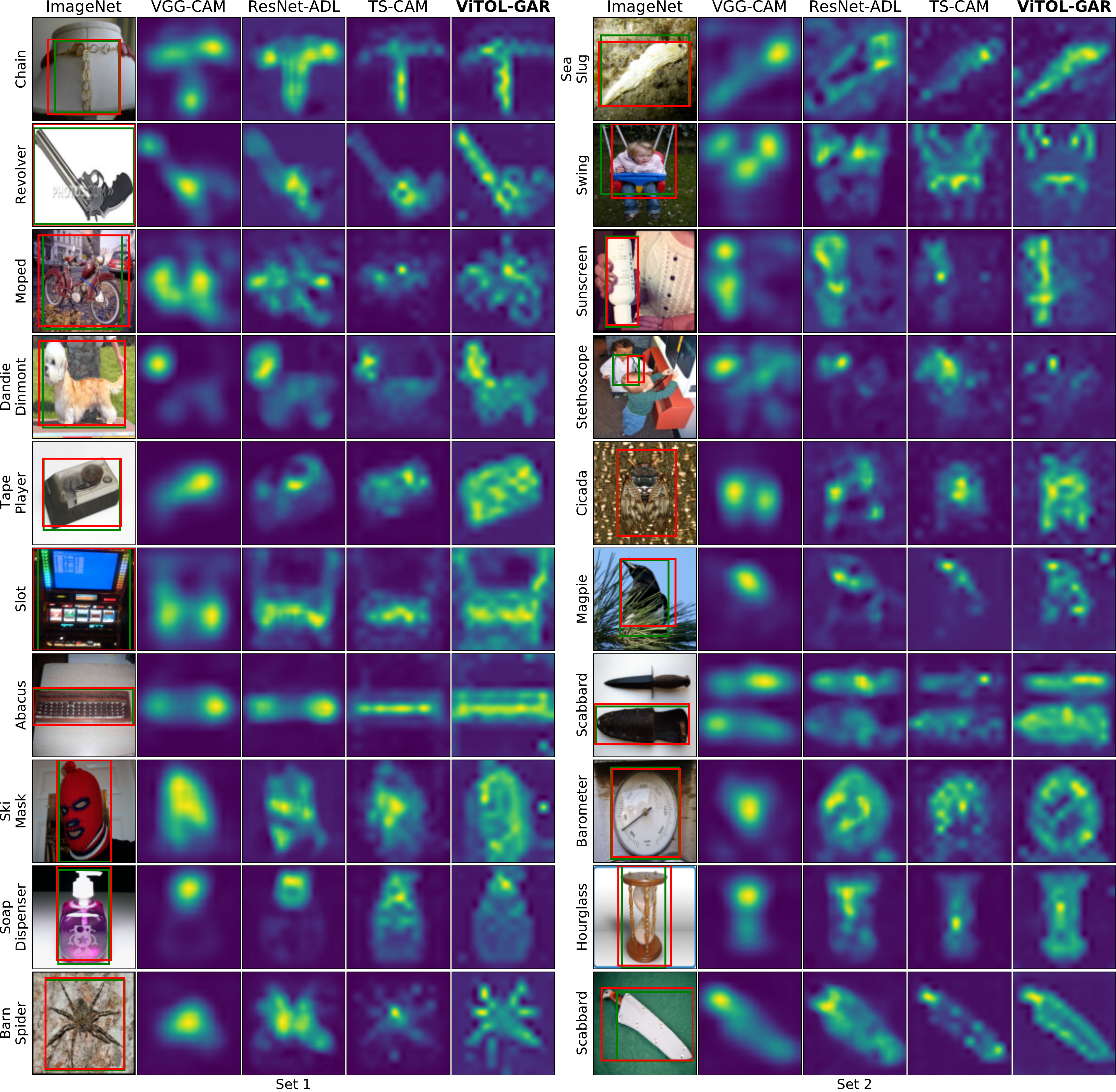}
\end{center}
  \caption{Comparison with state-of-the-art methods on 20 randomly sampled images from ImageNet validation split.}
\label{fig:SOTA}
\vspace{-1em}
\end{figure*}

Figure \ref{fig:SOTA} shows extensive comparison of localization maps for competing state-of-the-art (SOTA) methods against our proposed method ViTOL-GAR with 20 randomly sampled images from ImageNet dataset. We consider three models to draw comparisons: a) CAM \cite{zhou2016learning} based on VGG-16 backbone, b) ADL\cite{choe2019attention} trained on ResNet-50 backbone, and c) TS-CAM \cite{gao2021ts} trained on transformer DeiT backbone. 
We observe that ViTOL-GAR clearly outperforms competing approaches in its localization ability. Our method generates localization maps which (a) cover the entire region of the object, (b) are class dependent (e.g. Figure \ref{fig:SOTA} (Row Swing, Stethoscope) Set 2) ) and (c) invariant to background noise (e.g. Figure \ref{fig:SOTA} (Row Bam Spider Set 1, Cicada Set 2)).


\section{Ablation Study}
\begin{figure*}[t]
\begin{center}
\includegraphics[width=16cm, trim= 0 10cm 0 0, clip]{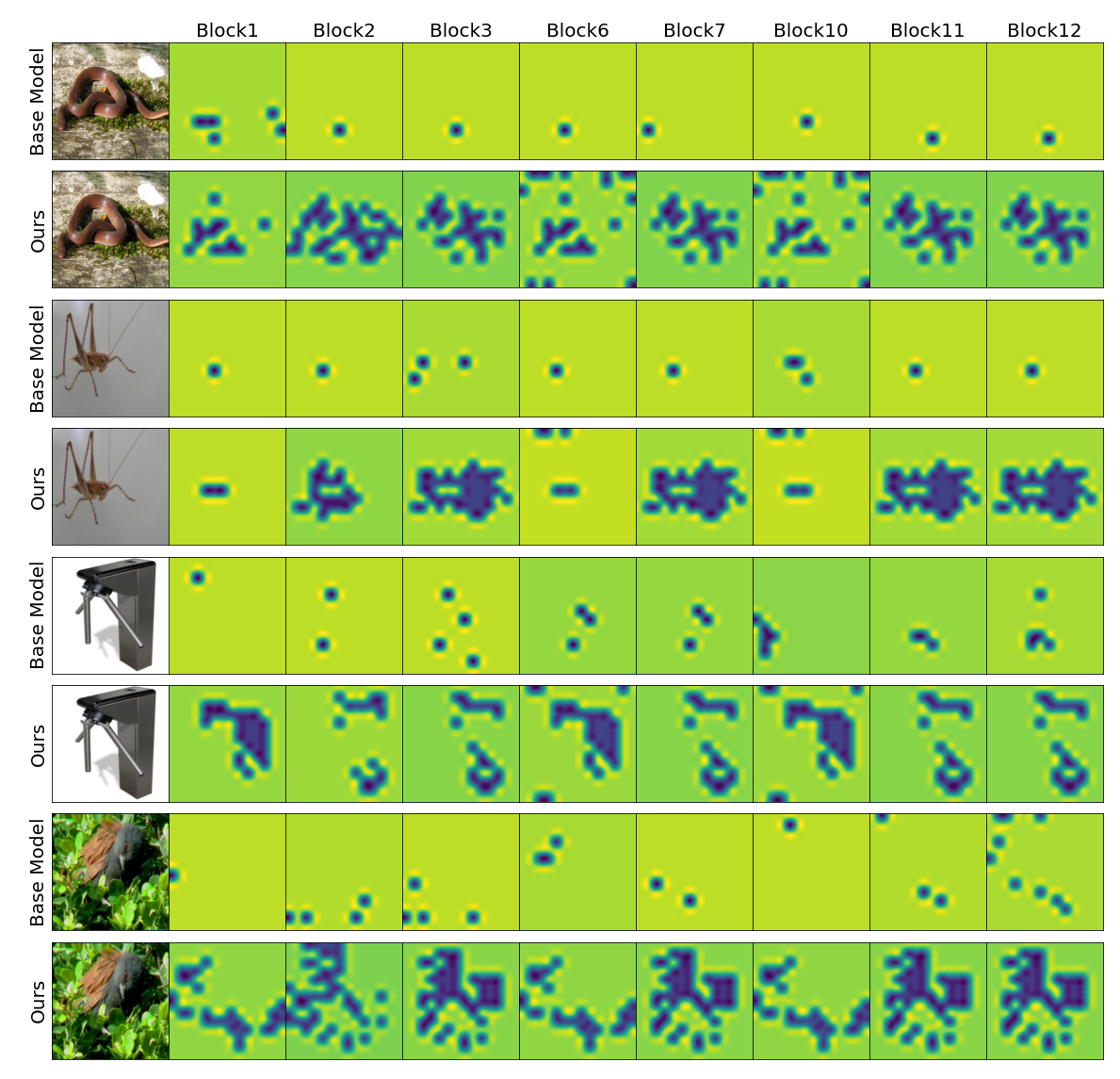}
\end{center}
  \caption{\textbf{Patch Drop masks:} Each alternate row shows the comparison of drop masks for some randomly sampled images in p-ADL layers of few encoder blocks. Base model refers to the pre-trained DeiT-B model. Ours is the DeiT-B model trained with p-ADL layers. Image samples are randomly chosen from ImageNet-1k validation split.}
\label{fig:Dropmask}
\end{figure*}

\subsection{GAR and LRP for different backbones}
Two attention map generation methods, GAR and LRP, have been used in our proposed approach. From the results in main manuscript, we observe that both these methods are effective in localizing the object. In this study, we aim to further understand differences in performance of these methods with different backbone architectures, namely, ViT-B, DeiT-S and DeiT-B. In Table \ref{table:GR_vs_LRP}, we observe that GAR performs consistently well across all backbones. Whereas LRP observes a drop of $1.59\%$, $0.78\%$ with ViT-B and DeiT-S backbones respectively on ImageNet dataset. 

GAR uses attention maps (attention matrix) of each layer and gradient maps corresponding to same attention map with respect to the desired class to calculate the final localization map. In addition to using these gradient maps, LRP also generates local relevance maps which is calculated based on \textit{local gradients} based on DTD principle in each layer using chain rule. This accumulation of local relevance across layers in LRP may negatively affect performance where embeddings do not effectively represent the regions belonging to the object. This can occur in cases of misclassification. Thus, features from a relatively weak classifier ViT-B may not be as representative of the correct class as compared to a strong classifier (DeiT-B). Intuitively, we believe this could be the reason for inconsistent performance of LRP with relatively weak classifier backbones such as ViT-B, DeiT-S as compared to DeiT-B. Moreover, we observed that LRP takes on average\textit{$\approx 2.26$x more time for running inference} over GAR while providing similar localization performance.

\begin{table}[t]
\begin{center}
\begin{tabular}{|c|ccc|}
\hline
       & \multicolumn{3}{c|}{ImageNet}                                                          \\ \hline
Method & \multicolumn{1}{c|}{ViT-B^{+}} & \multicolumn{1}{c|}{DeiT-S^{+}} & DeiT-B^{+} \\ \hline
GAR    & \multicolumn{1}{c|}{68.14}        & \multicolumn{1}{c|}{69.01}          & 69.17        \\ \hline
LRP    & \multicolumn{1}{c|}{66.55 }        & \multicolumn{1}{c|}{68.23}          & 70.47        \\ \hline
\end{tabular}
\end{center}
\caption{\textbf{GAR vs LRP for different transformer backbones}: MaxBoxAccv2 is shown for ImageNet. Superscript {(+)} denotes the architectures with p-ADL layer}
\label{table:GR_vs_LRP}
\end{table}

\subsection{Study of Patch Drop Mask}
In ViTOL, we use the p-ADL layer to enhance the localization capability. Patch drop mask and patch importance map are two key components of the p-ADL layer. Patch drop mask drops the most highlighted patches based on a drop threshold parameter ($\lambda$) and forces the model to look at less highlighted patches of the object of interest. However, only dropping the patches degrades the classification ability. Therefore, we use a patch importance map to retain the most highlighted patches to preserve the model's classification ability. We choose patch importance map or patch drop mask randomly at a chosen embedding drop rate ($\alpha$). 

In this study, we show that the p-ADL layer improves the attention in each encoder block progressively. In Figure \ref{fig:Dropmask}, we draw a comparison between the base transformer model against ViTOL by visualizing the drop mask of several encoder blocks. 
In our experiments, we use $\lambda=0.9$. Therefore, we drop the patches which have an attention value that is greater than $\lambda$ times the maximum attention value. 
Ideally, if the patch drop mask learns to drop the entire object region, it implies that each patch in the object region attains a value in the top $10\%$ of the values in the attention map. 
Our map generation mechanism uses this information to generate a self-attention map which uniformly attends to each patch covering the object.
In Figure \ref{fig:Dropmask} (Row 4), we observe that the Blocks 1 to 3 progressively drop the discriminative regions in the attention maps. However, from Blocks 4 and 5, the model potentially starts re-discovering the discriminative patches through the patch importance map. This behavior results in a model which is attentive to both, discriminative patches, as well as other features covering entire object.
Thereby resulting in a good quality localization map. 
Note, in Figure \ref{fig:Dropmask}, we only show those encoder blocks where the patch drop mask is chosen by the model.

\begin{table}[hb]
\begin{center}
\begin{tabular}{|c|c|c|}
\hline
$\alpha$ & $\lambda$ & MaxBoxAccv2 \\ \hline
0.5       & 0.9       & 66.37       \\ \hline
\textbf{0.75}      & \textbf{0.9}       & \textbf{72.42 }      \\ \hline
0.9       & 0.9       & 71.57       \\ \hline
0.5       & 0.8       & 68.2        \\ \hline
0.75      & 0.8       & 67.2        \\ \hline
0.9       & 0.8       & 70.2        \\ \hline
0.5       & 0.7       & 68.42       \\ \hline
0.75      & 0.7       & 68.81       \\ \hline
0.9       & 0.7       & 69.37       \\ \hline
\end{tabular}
\caption{Ablation of p-ADL parameters for experiments on the CUB dataset.}
\label{table:hyperparams}
\end{center}
\vspace{-1em}
\end{table}

In general, in Figure \ref{fig:Dropmask} we compare the patch drop mask behavior of the base model against our method. In the base model, the patch drop mask drops a very small portion of the entire object. This indicates that the entire object is not highlighted across different encoder blocks. This results in a poor localization performance. In contrast, the highlighted area in our method increases as the p-ADL layer drops the most discriminative patches and forces the model to focus on other parts of the object as well. In the encoder block $1$, the drop region does not cover most of the object of interest. However, as the p-ADL layer drops object specific regions across different encoder blocks, the network focuses on the other parts of the object. This enables the model to focus on the informative pacthes of the object. In some intermediate encoder blocks, patch importance mask is also selected which highlights the most discriminative region. And this discriminative portion is again dropped in subsequent encoder blocks. A similar pattern of highlighting and dropping region is observed across various query images of Figure \ref{fig:Dropmask}. Block $11$ and $12$ highlights most of the patches covering the object.

\subsection{p-ADL Hyperparameters Ablation}
In Table \ref{table:hyperparams}, we study the effect of changing different parameters of p-ADL layer. We vary the embedding drop rate $\alpha \in \{0.5, 0.75, 0.9\}$ and the patch drop threshold $\lambda \in \{0.7, 0.8, 0.9\}$ to show the trend in MaxBoxAccV2 localization metric. In our work, we choose $\alpha=0.75$ and $\lambda=0.9$ and use them consistently for all the experiments, as these result in the best localization score.

{\small
\bibliographystyle{ieee_fullname}
\bibliography{egbib}
}